# From Species to Cultivar: Soybean Cultivar Recognition using Multiscale Sliding Chord Matching of Leaf Images

Bin Wang[1], Yongsheng Gao[1,*], Xiaohan Yu[1,2], Xiaohui Yuan[2], Shengwu Xiong[2], Xianzhong Feng[3]

[1]School of Engineering, Griffith University, Australia
[2]School of Computer Science and Technology, Wuhan University of Technology, China
[3]Northeast Institute of Geography and Agroecology, Chinese Academy of Sciences, China

## Abstract

Leaf image recognition techniques have been actively researched for plant species identification. However it remains unclear whether leaf patterns can provide sufficient information for cultivar recognition. This paper reports the first attempt on soybean cultivar recognition from plant leaves which is not only a challenging research problem but also important for soybean cultivar evaluation, selection and production in agriculture. In this paper, we propose a novel multiscale sliding chord matching (MSCM) approach to extract leaf patterns that are distinctive for soybean cultivar identification. A chord is defined to slide along the contour for measuring the synchronised patterns of exterior shape and interior appearance of soybean leaf images. A multiscale sliding chord strategy is developed to extract features in a coarse-to-fine hierarchical order. A joint description that integrates the leaf descriptors from different parts of a soybean plant is proposed for further enhancing the discriminative power of cultivar description. We built a cultivar leaf image database, SoyCultivar, consisting of 1200 sample leaf images from 200 soybean cultivars for performance evaluation. Encouraging experimental results of the proposed method in comparison to the state-of-the-art leaf species recognition methods demonstrate the availability of cultivar information in soybean leaves and effectiveness of the proposed MSCM for soybean cultivar identification, which may advance the research in leaf recognition from species to cultivar.

---

* Corresponding author. Email: yongsheng.gao@griffith.edu.au, Tel: 61 7 37353652.

# 1. Introduction

Glycine max, commonly known as soybean, is a species of legume which has numerous uses such as acting as an important source of protein and oil for human and animal consumption. It is a major crop in the United States, Brazil, Argentina, India, and China, and has become one of the most widely consumed foods in the world due to its usefulness for human health and being easily cultivated [1]. Studies on soybean breeding, growth, development, and yield continue to be active research areas in both academia and agriculture industry. An important issue in soybean studies is the identification of soybean cultivar which plays a vital role in soybean cultivar evaluation, selection and production [2].

The applications of computer vision and image processing to leaf image analysis for plant species identification are being extensively studied. However, whether leaf images can also be used for cultivar identification remains an interesting problem yet to be investigated. The challenge in soybean cultivar identification from leaf images is that all the soybean cultivars belong to the same species of legume. The appearance of leaves from different cultivars are highly similar due to the fact that they are from plants of the same species, while the variation among different leaves of the same cultivar is relatively large (see Fig. 4). Hence, in contrast to identifying species using plant leaves that usually exhibit human identifiable pattern differences (see examples in species leaf image databases Leaf100 [9], MEW2012 [30], ICL [16] and Leafnap [15]), it is very difficult, even for human experts, to extract distinctive features from leaf images for soybean cultivar identification.

In this paper, we propose a novel multiscale sliding chord matching (MSCM) method for characterizing and recognizing soybean cultivars from their leaf images. When designing the proposed MSCM method, a chord is defined to slide along the leaf contour for measuring synchronised exterior shape features and interior appearance patterns of the soybean leaf image. Sliding chord measures in the form of signature functions are obtained at multiple scales, which provide discriminative powers in extracting shape and appearance information from coarsest global observations to finest details. A joint description by integrating the leaf descriptors from different parts of a soybean plant is proposed, which reveals in this research the usefulness and importance of complementary information contained in leaves at difference locations of a plant for classifying soybean cultivars. Experiments on classifying 200 soybean cultivars demonstrated encouraging capability of the proposed method, which achieved over 33% of higher accuracy than the state-of-the-art leaf image descriptors. To our knowledge, this is believed to be the first reported study (and cultivar leaf image database) on automated soybean cultivar identification by analysing leaf patterns.

The rest of this paper is organized as follows: In Section 2, the typical algorithms for leaf image analysis are reviewed. Section 3 presents the details of the proposed multiscale sliding chord matching method for soybean cultivar identification. A soybean cultivar leaf image database, SoyCultivar, is created and introduced in detail in Section 4. The experimental results of the proposed method and the state-of-art benchmarks are also presented in this section. Finally, conclusions are drawn in Section 5.

## 2. Related Work

Leaf visual features, such as leaf shape, texture, vein and colour, can be taken as significant cues for identifying plants. Different plants usually have their characteristic leaf shapes which play a key role for botanists to identify species [26]. There is a wealth of literature [43],[44],[45] describing leaf shapes for plant species identification. The leaf shape features can be extracted from the leaf boundary or from the area occupied by the leaf in the image plane. The former are generally named as contour based descriptors, while the later are classified as region based descriptors [31].

Leaf contours contain rich discriminative information about leaf margins, apexes and bases [8] which are usually taken as the main cues for species recognition. Thus, contour based descriptors have been developed for plant species recognition. Some of them focus on capturing the spatial distribution of contour points to characterize leaf shapes. Shape contexts [5] is a well-known contour based descriptor which treats the contour as a set of landmarks. Each landmark is taken as a reference point and the distribution of the relative positions of the remaining ones are computed to form a log-polar histogram. In order to achieve invariance to articulation, Ling et al. [6] defined the distance between two landmarks as the length of the shortest path between them within the shape silhouette and proposed an inner-distance shape contexts (IDSC) method for robust shape classification. This approach has been effectively applied to a real system for plant leaf identification [7]. Hu et al. [16] proposed an invariant contour descriptor, multiscale distance matrix (MDM), for fast leaf recognition in which the relationships between each pair of landmarks of the leaf contour are finely characterized by a variety of distances including Euclidean distance and inner-distance. Backes et al. [10] modelled the shape contour into a small-world complex network and used the degree and joint degree measurements in a dynamic evolution network to yield a set of shape descriptors. It is not only invariant to scale and rotation changes, but also tolerant to shape deformation and noise on contour.

Recently, Zhao et al. [8] proposed a pattern counting method for the classification of both simple leaves and compound leaves. A novel feature, independent-IDSC, is developed for better capturing global and local information of a leaf, such as the overall shape, margin type of a simple leaf, and leaflet details of a compound leaf. The number of their patterns are counted and the resultant histogram is taken as the descriptor of the entire leaf. Hierarchical string cuts (HSC) [9] characterizes a contour segment using the spatial distribution information of the contour points relative to its string that cuts the segment off the whole contour. This method is very fast and suitable for large shape database retrieval with a highly competitive accuracy.

Some methods regard the leaf contour as a curve and measure its curvature property for leaf recognition. The curvature scale space (CSS) [32] is a contour descriptor suggested by MEPG-7 [33]. The CSS image of a shape is a multiscale organization of its inflection points as it is smoothed and contains several arch shape contours with each being related to a concavity or a convexity of the shape. In [33], a generalization of the CSS representation is proposed to recognize 2-D contours with self-intersections and is successfully used for classifying Chrysanthemum leaves. Chen et al. [34] proposed a velocity representation, derived from the first derivative of the distance vector which is available by calculating the distance of each contour point to the centre of the contour, for the classification of weed leaf images. This method was shown to be easier and faster at finding the contour shape characteristics than the CSS.

Measuring the curvature property of a leaf contour usually requires differential operations. However the differential techniques have the inherent sensitivity to noise [35]. Instead of using differential operations, Manay et al. [35] proposed using integral operations to derive a class of functionals for robust curvature measurement. The obtained integral invariants preserve the desirable properties of the differential measures, such as allowing matching under occlusions and uniqueness of representation, yet more robust to noise. Kumar et al. [15] used integral measures to compute functions of the curvature for each boundary point and created histograms of curvature values over different scales. They developed a mobile app for identifying plant species using this technique. Alternative measures of curvature, such as arch height [17] and triangle area [18, 19], have also been proposed for leaf image identification and achieved attractive results.

Region based leaf descriptors focus on the whole leaf region for charactering leaf patterns. Image moments are a widely applied category of descriptors for object recognition. Wang et al. [21] classified leaf images using a combination of Hu geometric moments and Zernike moments. Horaisová and Kukal [20] treated the leaf shape as a binary image pattern in which 2D Fourier power spectrums are extracted for translation, scaling, rotation and mirroring invariant leaf classification. Considering that significant curvature points are hard to find from a leaf contour, Lee et al. [22] suggested that some region based features, such as aspect ratio, compactness, centroid, and horizontal/vertical projections, are more reliable for leaf image identification. Recently, integral transform methods like Radon transform based descriptors [46], [47], structure integral transform [48], and multiscale integral invariants [49] are also developed for extracting geometrical features from the shape region.

Apart from the leaf shape characteristics, texture patterns of a leaf image also carry discriminative information which can be taken as a significant cue for classifying plant species. In [25], [36], [37], [39], [40], the popular texture descriptors such as Gabor filer, fractal dimensions and gray-level co-occurrence matrix have been applied to leaf image analysis. Ling et al. [6] extended the IDSC to the shortest path texture context (SPTC) by measuring the distributions of weighted relative orientations of local intensity gradient of the shape image, where the weights are derived from the gradient magnitudes. The experimental results in their paper show that the SPTC achieves a 1.2% accuracy increase over the IDSC on the Swedish species leaf database. Veins are known as a unique structure of leaves that transport liquids and nutrients to leaf cells. Several typical vein patterns such as parallel, palmate and pinnate can be found in many plant leaf species. Larese et al. [23] proposed an automatic procedure of classifying three legume species, soybean, red and white beans, based only on the analysis of the leaf vein morphological features. Park et al. [41] extracted structure features for categorizing venation patterns of leaves. More studies associated with plant species identification can be found in [26].

Inspired by the bag-of-words model, Wang et al. [50] proposed to decompose shapes into contour fragments and describe them by shape contexts [5] that are treated as shape codes. A set of shape codes randomly selected from the training shapes are then used to learn a shape codebook. Based on the codebook, a shape can be encoded for robust shape classification. Lee et al. [51] studied the use of deep learning to harvest discriminatory features from leaf images and apply them for plant species identification. They used a large number of training samples, 40-60 samples per class, for training their CNN model.

Although extensive studies have been conducted on plant species identification from leaf images, very few works are done on leaf analysis for cultivar identification. We made the first attempt to investigate the possibility of identifying soybean cultivars from their leaf images. A leaf image database of 200 soybean cultivars, the first of its kind, has also been created for algorithm evaluation on cultivar leaf recognition.

## 3. The Proposed Method

In this work, we propose a leaf image description modelled by a set of chord integrals. They play different roles of characterizing a leaf image and provide an overall consideration for extracting features across leaf contour, region and appearance. Although chord is a geometrical primitive usually used to characterize shape properties, our model employs a novel scheme to measure the texture information along the chord which makes the descriptor effectively fuse the shape and texture features for more accurate object description. The proposed method comprehensively characterizes the leaf image from the perspectives of the shape and texture at multiple scales, which makes different types of features complementary with each other.

Mathematically, a gray-level leaf image can be represented as a 2D function $g(x,y)$ that describes the intensity of a pixel at Cartesian coordinates $x$ and $y$ in the image plane. The outer contour $\Omega$ of the leaf shape can be represented in an arc-length parameterization form [27]: $z(t) = (\bar{x}(t), \bar{y}(t)), t \in [0,1)$, where $(\bar{x}(t), \bar{y}(t)) \in \Omega$. Note that the perimeter length of the contour has been normalized to 1 unit length. Since $\Omega$ is a closed contour, the function $z(t)$ is a periodic function and its period is 1. Thus, we have $z(t+1) = z(t)$ and $z(t-1) = z(t)$. Let D denote the region in the image plane enclosed by the contour $\Omega$. The leaf shape can thus be represented by a binary function $f(x,y) = 1$ if $(x,y) \in D$, and 0 otherwise. In this section, $g(x,y)$, $z(t)$, and $f(x,y)$ that is image intensity, contour and shape of a leaf image will be used in formulating the detailed characteristics of soybean leaves for the purpose of cultivar classification.

### 3.1. Chord Measures

Given a point $z(t) = (\bar{x}(t), \bar{y}(t))$ on the contour $\Omega$ and an arc length $r \in (0,1)$, we can create a chord $L_{t,r}$ (named as $r$-arc-length chord) that connects the arc's start point $z(t)$ and end point $z(t+r)$. Let $l$ denote the length of the chord $L_{t,r}$. For a point $p$ on the chord, its location (i.e., coordinates) can be determined by

$$c(t, r, \tau) = (\bar{x}(t) + \tau \cos \theta, \bar{y}(t) + \tau \sin \theta), \qquad (1)$$

where $\tau \in [0, l]$ is the distance from the start point $z(t)$ to the point $p$ along the chord $L_{t,r}$. $\theta$ is the orientation of the chord determined by $z(t)$ and $z(t+r)$, and we have

$$\cos \theta = \frac{\bar{x}(t+r) - \bar{x}(t)}{l}, \qquad \sin \theta = \frac{\bar{y}(t+r) - \bar{y}(t)}{l}. \qquad (2)$$

Firstly, we use the chord $L_{t,r}$ to measure the geometrical characteristics of the leaf shape via calculating the integral of the binary shape function $f(x, y)$ over the chord $L_{t,r}$ as

$$\eta(t,r) = \int_0^l f(c(t,r,\tau))d\tau. \qquad (3)$$

It calculates the length of the part of the chord $L_{t,r}$ that passes through the leaf shape region D to depict the morphological properties of the leaf. Given an arc length $r$, if the leaf shape is a convex, we have a varying $\eta(t,r) = l$ due to the change of $l$ when the start point $z(t)$ moves along the contour $\Omega$. If the leaf shape is a concave (see Fig. 1), we have a varying $\eta(t,r)$ ranging from 0 to $l$ and $l$ also changes at the same time when the start point $z(t)$ moves along the contour $\Omega$.

Secondly, the structure information of the leaf contour is also encoded along the chord $L_{t,r}$ as

$$h(t,r) = \frac{1}{r}\int_0^r d(t,r,s)ds, \qquad (4)$$

where $s \in (0, r)$ and $d(t, r, s)$ denotes the perpendicular distance from the contour point $z(t + s)$ to the chord $L_{t,r}$, which can be calculated by

$$d(t,r,s) = \frac{1}{l}\left|\det\begin{pmatrix} \bar{x}(t) & \bar{y}(t) & 1 \\ \bar{x}(t+s) & \bar{y}(t+s) & 1 \\ \bar{x}(t+r) & \bar{y}(t+r) & 1 \end{pmatrix}\right|, \qquad (5)$$

where $\det(\cdot)$ denotes the determinant of a matrix. The signature $h(t, r)$ depicts how much the $r$-length arc from the start point $z(t)$ to the end point $z(t + r)$ on the contour deviates away from its chord $L_{t,r}$ when the start point $z(t)$ moves around the contour $\Omega$. When $h(t, r)$ approaches a value of 0, it means that start point $z(t)$ trespasses a structurally flat part of arc on the contour.

Thirdly, the appearance, that is the intensity information, of the leaf is considered in our design. We capture the appearance information in the shape region D along the chord $L_{t,r}$:

$$\mu(t,r) = \frac{1}{\eta(t,r)} \int_0^l g(c(t,r,\tau)) f(c(t,r,\tau)) d\tau, \tag{6}$$

$$\sigma(t,r) = \sqrt{\frac{1}{\eta(t,r)} \int_0^l \big(g(c(t,r,\tau)) - \mu(t,r)\big)^2 f(c(t,r,\tau)) d\tau}, \tag{7}$$

where the binary shape term $f(c(t,r,\tau))$ is used to constrain the integral of intensity function $g(c(t,r,\tau))$ to be only over the part of the chord within the shape region D. The $\mu(t,r)$ depicts the intensity strength of the pixels inside the shape region D along the chord $L_{t,r}$, while $\sigma(t,r)$ describes the extent of the intensity changes of pixels in the shape region D along the chord $L_{t,r}$. Their values vary when the start point $z(t)$ moves along the contour Ω, which encodes the intensity information of the leaf into distinctive signature curves. When $\eta(t,r) = 0$, that is the chord $L_{t,r}$ falls completely outside the shape region D, $\mu(t,r)$ and $\sigma(t,r)$ are directly set to 0 in our implementation.

Fig. 1 visually illustrates the extraction process of the four chord measures, $\eta$, $h$, $\mu$, and $\sigma$, from a soybean leaf image. The geometrical and structural shape features of a leaf are characterized by $\eta$ and $h$, while the appearance features are characterized by $\mu$ and $\sigma$.

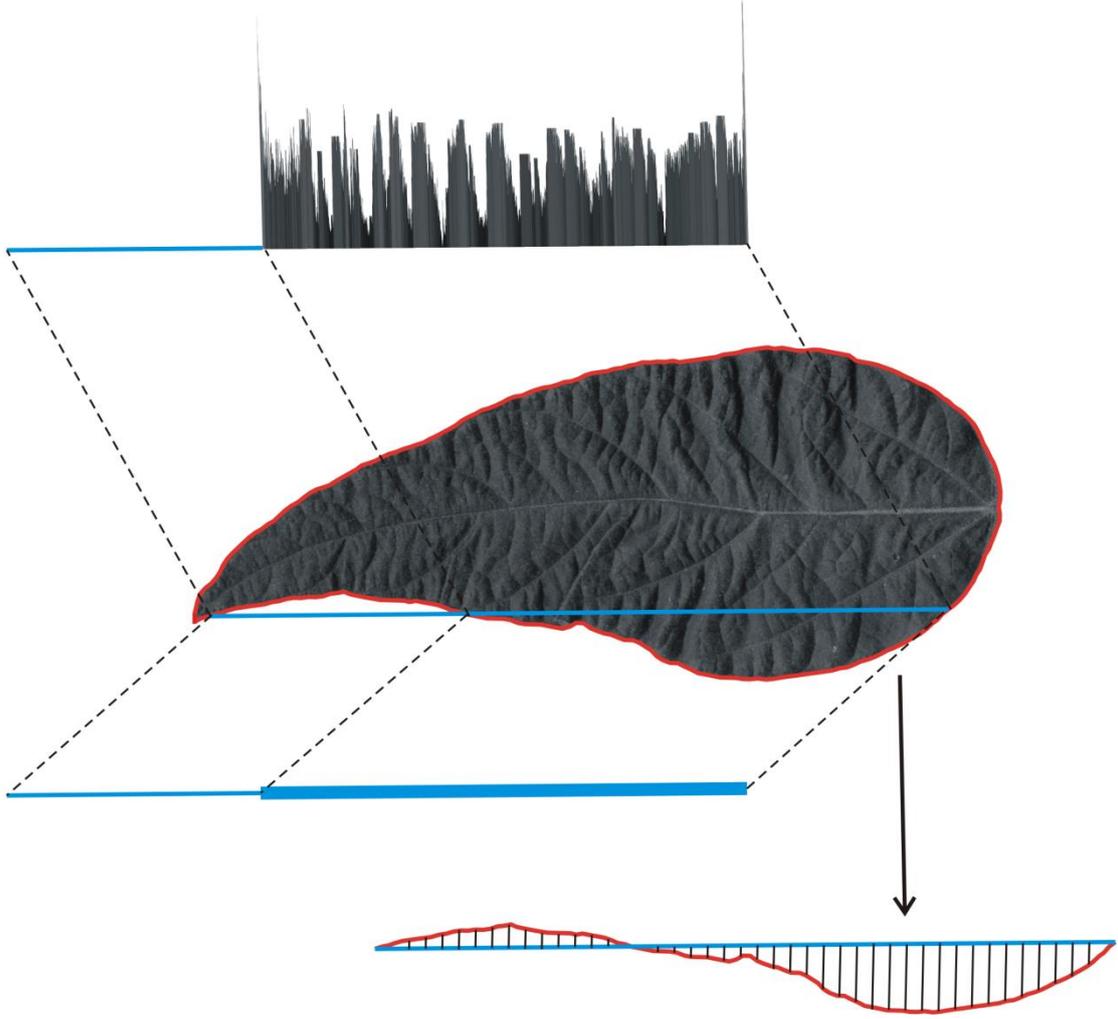

Fig. 1. A visual illustration of extracting the four chord measures, $\eta$, $h$, $\mu$, and $\sigma$, for a soybean leaf. The chord is marked in blue colour, and the contour is marked in red colour. $\eta$, $h$ are displayed below the leaf, while $\mu$, $\sigma$ are visualised above the leaf.

## 3.2. Multiscale Sliding Chord Matching

A new multiscale sliding chord matching method is proposed in this section. Give a $r$-arc-length chord $L_{t,r}$, by sliding its start point $z(t)$ around the contour, i.e., varying the value of the parameter $t$ from 0 to 1, the leaf region traversed by the chord $L_{t,r}$ can be described by four signature functions $\eta_r(t)$, $h_r(t)$, $\mu_r(t)$, and $\sigma_r(t)$, for $t \in [0,1)$.

In our proposed method, the parameter $r$ naturally plays a role of scale for constructing a scale-space to effectively describe the synchronised shape and appearance features in multiple scales. Considering that $z(t + 1) = z(t)$, the farthest location where the start point $z(t)$ of a chord $L_{t,r}$ can slide to along the contour is $z\left(t + \frac{1}{2}\right)$. Thus, the scale parameter $r$ varies within the range of $(0, \frac{1}{2}]$, and the largest scale is $r = 2^{-1}$. Instead of using the popular linear scale arrangement [16][18][19], we employ a log-scale space $r \in \{2^{-1}, \cdots, 2^{-K}\}$ similar to the log-polar space used in [5]. $K$ is the total number of scales. This can make the descriptor more sensitive to the nearby neighbourhood of the contour point $z(t)$ than to those farther away. For the largest scale where the chord $L_{t,r}$ with $r = 2^{-1}$ slides along the contour, the whole leaf region is scanned in a globally coarse description manner (see the top leaf image in Fig. 2). With the decrease of $r$, the chord $L_{t,r}$ becomes shorter and the area scanned by sliding the chord progressively moves towards the peripheral region of the leaf which depicts finer details of the leaf shape (see the bottom leaf images in Fig. 2). Together, $4K$ signature functions of $\eta_r(t)$, $h_r(t)$, $\mu_r(t)$, and $\sigma_r(t)$ for $r = 2^{-1}, \cdots, 2^{-K}$ can be created to hierarchically describe both the shape and appearance of a leaf in a coarse to fine and global to peripheral manner. Fig. 2 shows an example of the proposed multiscale sliding chord process on a soybean leaf that generated 12 signature functions for three scales $r = 2^{-1}, \cdots, 2^{-3}$.

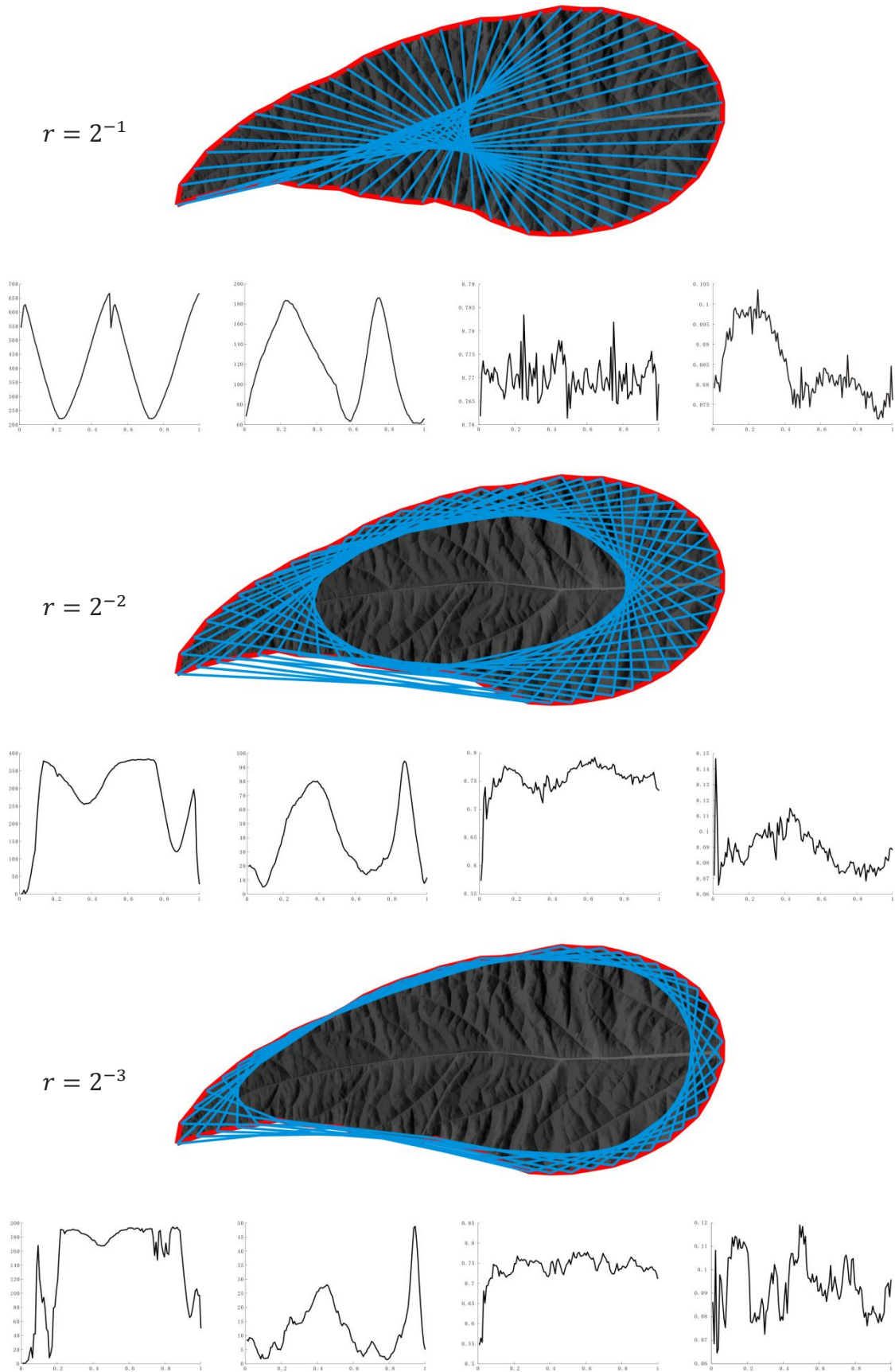

Fig. 2. An example of the proposed sliding chord process on a soybean leaf at scales $r = 2^{-1}, \cdots, 2^{-3}$. The curves under each leaf image are the generated signature functions of $\eta$, $h$, $\mu$, and $\sigma$ respectively (from left to right).

## 3.3. Invariance and Normalization

A desirable leaf descriptor should not only have a high discriminative ability, but also be invariant to rotation, scaling, and translation. Because translating or rotating a leaf in the image plane does not change the position of the chord $L_{t,r}$ relative to its corresponding arc and the shape region D, the four measures $\eta(t,r)$, $\mu(t,r)$, $\sigma(t,r)$ and $h(t,r)$ are thus intrinsically invariant to translation and rotation.

When a leaf in the image plane is subjected to a scaling transform, its associated functions $g(x,y)$, $f(x,y)$, and $z(t)$ will become $g'(x,y) = g(u,v) = g(x/a, y/a)$, $f'(x,y) = f(x/a, y/a)$ and $z'(t) = a \cdot z(t)$, where $a > 0$ is the scale factor and $x = au, y = av$. Then the two end points $z(t)$ and $z(t+r)$ of the chord $L_{t,r}$ are scaled to $z'(t) = a \cdot z(t)$ and $z'(t+r) = a \cdot z(t+r)$ and the length $l$ of the chord $L_{t,r}$ becomes $\cdot l$. For a point $p$ on the chord $L_{t,r}$, its distance $\tau$ from the start point $z(t)$ of the chord is scaled to $\tau' = a \cdot \tau$. Because the parameters $t, r$ have been normalized, the location (i.e., coordinates) of the point $p$ (see Eq. 1) after scaling transform becomes

$$c'(t, r, \tau') = (a \cdot \bar{x}(t) + \tau' \cos\theta, a \cdot \bar{y}(t) + \tau' \sin\theta)$$
$$= (a \cdot \bar{x}(t) + a \cdot \tau \cos\theta, a \cdot \bar{y}(t) + a \cdot \tau \sin\theta)$$
$$= a \cdot c(t, r, \tau). \tag{8}$$

Then, we have

$$\eta'(t,r) = a \cdot \eta(t,r), \tag{9}$$
$$h'(t,r) = a \cdot h(t,r), \tag{10}$$
$$\mu'(t,r) = \mu(t,r), \tag{11}$$

and

$$\sigma'(t,r) = \sigma(t,r). \tag{12}$$

The proofs of above equations are presented in Appendix A.

Eq. (11) and Eq. (12) indicate that $\mu(t,r)$ and $\sigma(t,r)$ are invariant to scaling transform. While Eq. (9) and Eq. (10) show that measures $\eta(t,r)$ and $h(t,r)$ are scale variant by the scale factor of $a$. Therefore, for the 1D signature functions $\eta_r(t)$ and $h_r(t)$, $r = 2^{-1}, \cdots, 2^{-K}$, we can easily normalize them into scale invariant forms as $\eta_r(t)/\max_{0 \leq t < 1}\{\eta_r(t)\}$ and $h_r(t)/\max_{0 \leq t < 1}\{h_r(t)\}$, $r = 2^{-1}, \cdots, 2^{-K}$, respectively.

Since the parameterized contour curve function $z(t)$ changes when the starting point shifts on the contour, all the signatures $\eta_r(t), h_r(t), \mu_r(t)$, and $\sigma_r(t)$ varies according to the starting point of the contour $\Omega$. Consider that selecting different contour points as the starting point only produces a phase shift of the function $z(t)$ along $t$ axis, which results in all the signatures $\eta_r(t), h_r(t), \mu_r(t)$, and $\sigma_r(t)$ having the same phase shift as $z(t)$. Here we use the magnitudes of their Fourier transforms to remove the effect of such phase shifts. Let $\tilde{\eta}_r(\omega) = \left\|\int_0^1 \eta_r(t)e^{-2\pi\omega j t}dt\right\|$, $\tilde{h}_r(\omega) = \left\|\int_0^1 h_r(t)e^{-2\pi\omega j t}dt\right\|$, $\tilde{\mu}_r(\omega) = \left\|\int_0^1 \mu_r(t)e^{-2\pi\omega j t}dt\right\|$, and $\tilde{\sigma}_r(\omega) = \left\|\int_0^1 \sigma_r(t)e^{-2\pi\omega j t}dt\right\|$, where $j^2 = -1$ and $\|\cdot\|$ denotes the magnitude of a complex number. We take the first $C$ Fourier coefficients for each signature to construct compact and invariant signatures $\tilde{\eta}_r(\omega), \tilde{h}_r(\omega), \tilde{\mu}_r(\omega), \tilde{\sigma}_r(\omega)$ for $\omega = 0, \cdots, C-1$ and $r = 2^{-1}, \cdots, 2^{-K}$ (In our experiments, we empirically set $C = 7$ and $K = 7$).

Finally, a soybean leaf descriptor $\Phi$, a vector of dimension $4C \cdot K$, is designed as

$$\Phi = \{\bar{\eta}_r(\omega), \bar{h}_r(\omega), \bar{\mu}_r(\omega), \bar{\sigma}_r(\omega)\}_{\omega=0,\cdots,C-1, r=2^{-1},\cdots,2^{-K}} \quad (13)$$

where $\bar{\eta}_r(\omega) = \tilde{\eta}_r(\omega)/\dot{\eta}$, $\bar{h}_r(\omega) = \tilde{h}_r(\omega)/\dot{h}$, $\bar{\mu}_r(\omega) = \tilde{\mu}_r(\omega)/\dot{\mu}$, and $\bar{\sigma}_r(\omega) = \tilde{\sigma}_r(\omega)/\dot{\sigma}$. To balance their contributions, each compact and invariant signature ($\tilde{\eta}_r(\omega), \tilde{h}_r(\omega), \tilde{\mu}_r(\omega), \tilde{\sigma}_r(\omega)$) is normalized by the average value of its $0^{th}$-order Fourier coefficients ($\omega = 0$) obtained from the training data (in our experiment, they are obtained from the model leaves only) as

$$\dot{\eta} = \frac{1}{N \cdot K} \sum_{i=1}^{N} \sum_{r=2^{-1}}^{2^{-K}} \tilde{\eta}_r^i(0), \qquad \dot{h} = \frac{1}{N \cdot K} \sum_{i=1}^{N} \sum_{r=2^{-1}}^{2^{-K}} \tilde{h}_r^i(0),$$

$$\dot{\mu} = \frac{1}{N \cdot K} \sum_{i=1}^{N} \sum_{r=2^{-1}}^{2^{-K}} \tilde{\mu}_r^i(0), \qquad \dot{\sigma} = \frac{1}{N \cdot K} \sum_{i=1}^{N} \sum_{r=2^{-1}}^{2^{-K}} \tilde{\sigma}_r^i(0). \tag{14}$$

where $N$ is the number of training leaf images.

## 3.4. Joint Description and Similarity Comparison

The leaves from different parts of a soybean plant usually have different shapes and appearance (see an example in Fig. 3), which indicate that the leaves from different parts of a plant may provide different but complementary cues for cultivar classification of soybean plants. This observation motivates us to propose a joint description by integrating the leaf descriptors from different parts of a soybean plant for enhancing the discriminative power of cultivar description.

Let $\Phi^U$, $\Phi^M$, $\Phi^L$ denote the descriptors of the leaves taken from the upper part, the middle part and the lower part of a soybean plant. We integrate them to form a joint leaf descriptor $\psi = \{\Phi^U, \Phi^M, \Phi^L\}$ for characterizing the soybean plant. Assume that $\psi_A = \{\Phi_A^U, \Phi_A^M, \Phi_A^L\}$ and $\psi_B = \{\Phi_B^U, \Phi_B^M, \Phi_B^L\}$ are two joint leaf descriptors for soybean plant A and soybean plant B, respectively. Their dissimilarity can be measured by

$$D(A,B) = |\Phi_A^U - \Phi_B^U| + |\Phi_A^M - \Phi_B^M| + |\Phi_A^L - \Phi_B^L|, \tag{15}$$

where $|\cdot|$ is $L_1$-norm and $|\Phi_A - \Phi_B|$ is defined by

$$|\Phi_A - \Phi_B| = \sum_{\omega=0}^{C-1} \sum_{r=2^{-1}}^{2^{-K}} W\big(|\bar{\eta}_r^A(\omega) - \bar{\eta}_r^B(\omega)| + |\bar{h}_r^A(\omega) - \bar{h}_r^B(\omega)|\big)$$

$$+ (1-W) \cdot (|\bar{\mu}_r^A(\omega) - \bar{\mu}_r^B(\omega)| + |\bar{\sigma}_r^A(\omega) - \bar{\sigma}_r^B(\omega)|). \tag{16}$$

$W \in [0,1]$ is a weighting factor used for adjusting the contributions of the shape features (i.e., $\bar{\eta}$ and $\bar{h}$) and the appearance features (i.e., $\bar{\mu}$ and $\bar{\sigma}$) for classifying soybean cultivars.

When $W = 1$, only the shape features are used for dissimilarity measurement, while $W = 0$ indicates that only the appearance features are used to measure the dissimilarity of two cultivar plants.

## 4. Experimental Results and Discussions

Some of the experienced breeders feel that there seems to be some relationship between soybean cultivars and their leave patterns that may be useful for soybean cultivar classification, which remains a hypothesis to be tested. In this research, we built a soybean cultivar leaf database, SoyCultivar, by collecting the leaf images from soybean plants of different cultivars that grow in Jilin Provence known as a major soybean production area in China. The SoyCultivar database contains 1200 leaves collected from plants of 200 soybean cultivars. For each cultivar, we randomly collected two leaves from the upper part, two leaves from the middle part, and two leaves from the lower part of the plants. Both the reflective and transparent scans of the front and back sides of the leaves are obtained using an EPSON V800 scanner with a resolution of 600 DPI and 24 bit true color setting. Four images are scanned for each leaf (one reflective image of the front side, one reflective image of the back side, one transparent image of the front side, and one transparent image of the back side).

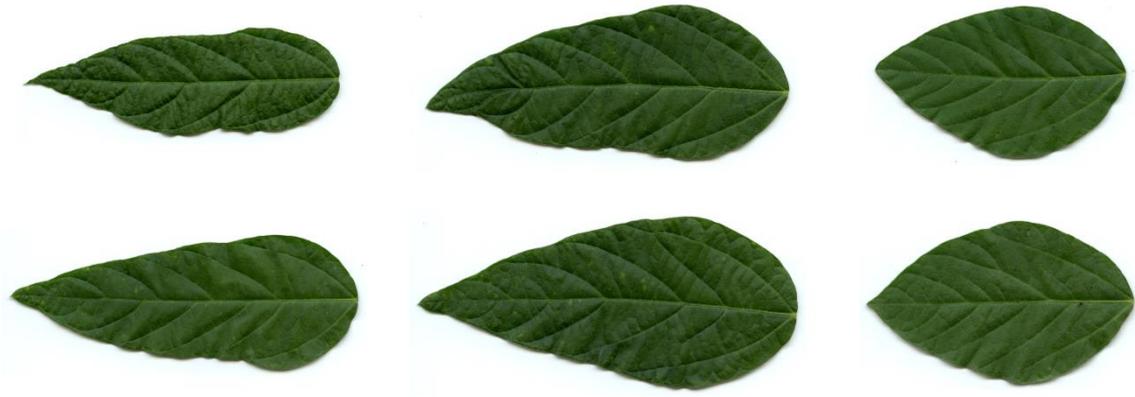

Fig. 3. An example of all six leaves from soybean plants of one cultivar. The leaves listed in the columns from left to right are taken from the upper, middle and lower parts of the plants of a soybean cultivar.

In this study, the reflective images captured from the front sides of leaves of all the 200 soybean cultivars are used to evaluate the effectiveness of the proposed method. There are a total of $200 \times 2 \times 3 = 1200$ leaf images used in our experiments. Fig. 3 shows an example of all six leaves taken from the plants of one cultivar. The leaves taken from the soybean plants of the first 50 cultivars are shown in Fig. 4 as examples of the SoyCultivar database. It shows 300 leaf images of the first 50 cultivars, which are displayed in 10 tables with 5 cultivars in each table. Each row of a table displays all the six leaves taken from the plants of the same cultivar. Each column shows three leaves (from left to right) which are taken from the upper, middle and lower parts of the plants of one soybean cultivar, respectively.

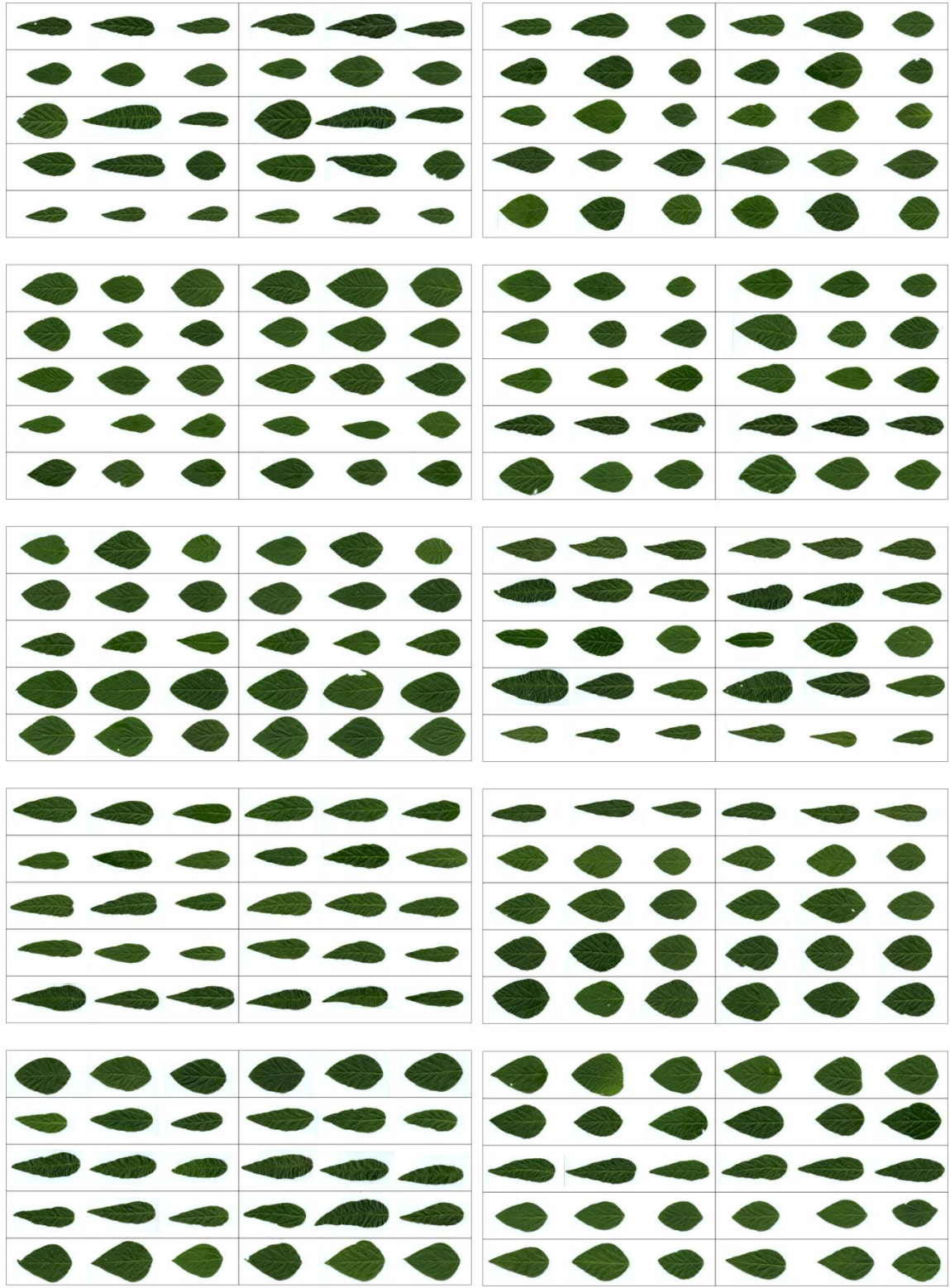

Fig. 4. The front side reflective images for leaves of the first 50 soybean cultivars in the SoyCultivar database. They are displayed in 10 tables and each table contains 5 cultivars. Each row of a table displays all the six leaves from the same cultivar. Each column contains three leaves (from left to right) which are taken from the upper, middle and lower parts of the plants of one soybean cultivar, respectively.

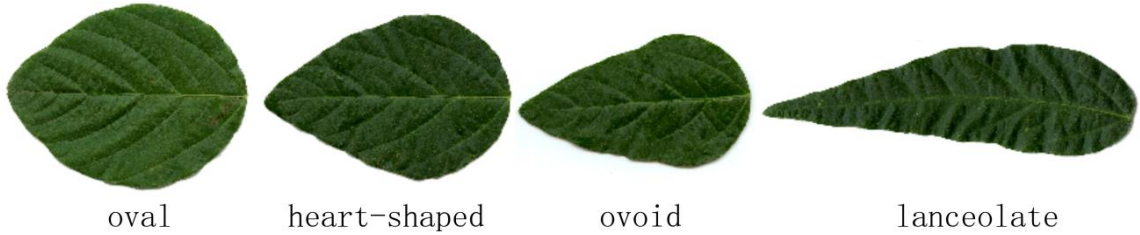

Fig. 5. Typical shapes of soybean leaves in the database.

It is noticed that the soybean leaves in the SoyCultivar database can be broadly categorised into oval, heart-shaped, ovoid and lanceolate shapes (see examples in Fig. 5). For some soybean cultivars, their leaves from different parts of the plants exhibit a quite diverse appearance in completely different shapes, while for other soybean cultivars, their leaves from different parts of the plants have the same shape. However, compared to the species leaf image databases Leaf100[9], MEW2012[30], ICL[16], the leaves in the SoyCultivar database are highly similar due to the fact that they all belong to the same species, making it a new and challenging dataset for the pattern recognition research community.

## 4.1. Evaluation Protocol

For each soybean cultivar, there are two samples in the SoyCultivar database. The first samples of all the 200 cultivars are used to construct the model set, while the second samples of the 200 cultivars are used to construct the test set. There are 600 images of 200 cultivars in each set. Each sample in the test set is matched against all the samples in the model set, which results in 200 matching tests. Then, the roles of these two sets are interchanged, i.e. the first samples are used as test samples and the second samples are used as models, which makes another 200 matching tests. In total, there are 400 matching tests. A correct identification is counted, when the best matched sample from the model set is from the same cultivar of the test sample. The identification accuracy is calculated by $n/400$, where $n$ is the number of correct identification tests.

## 4.2. Results and Discussion

To examine if the soybean plant leaves contain discriminative pattern information for identifying cultivars, experiments are conducted to evaluate the performance of the proposed Multiscale Sliding Chord Matching (MSCM) method and compared with those of the state-of-the-art leaf identification benchmark methods. They are: (1) Hierarchical String Cuts (HSC) [9], (2) two versions of Multiscale Distance Matrix [16], MDM-CD-RM and MDM-ID-RA, (3) Inner Distance Shape Contexts (IDSC) [6], (4) Shortest Path Texture Context (SPTC) [6], (5) Bag of Contour Fragments (BCF) [50], and (6) Structure Integral Transform (SIT) [48]. Among them, HSC, MDM, IDSC and BCF are contour based methods, SIT is a region based method, and SPTC is a method that can encode both shape and texture features. For the BCF method, its original version[1] uses the SVM as the classifier. Because all the other competing methods use the nearest neighbour (1NN) classifier, the BCF using $L_1$-norm distance 1NN classifier is also implemented as a benchmark for fair comparison. Score level fusion is used when implementing the benchmark algorithms for joint matching of the leaves from the upper, middle and lower parts.

Table 1. The identification accuracy (%) of the proposed method together with those of the benchmarks on the SoyCultivar database.

| Algorithm | Upper Part | Middle Part | Lower Part | Joint Matching |
|---|---|---|---|---|
| HSC [9] | 14.50 | 13.25 | 13.25 | 34.50 |
| MDM-CD-RM [16] | 11.50 | 9.00 | 12.25 | 32.00 |
| MDM-ID-RA [16] | 12.25 | 9.50 | 9.25 | 32.25 |
| IDSC [6] | 11.00 | 13.25 | 7.00 | 33.25 |
| SPTC [6] | 12.75 | 14.50 | 11.25 | 40.25 |
| BCF+SVM [50] | 14.25 | 9.25 | 9.50 | 34.25 |
| BCF [50]+1NN | 12.75 | 9.70 | 7.75 | 29.00 |
| SIT [48] | 15.00 | 13.50 | 9.25 | 40.00 |
| **Proposed MSCM** | **30.50** | **23.00** | **26.25** | **67.50** |

---

[1] The source code downloaded from the authors' website.

The identification accuracies of the proposed MSCM approach and the benchmark approaches are summarised in Table 1. It is very encouraging to observe that the proposed method achieved an exciting accuracy of 67.50% on such a challenging cultivar leaf recognition task. It is 33.00%, 35.50%, 35.25%, 34.25%, 27.25%, 33.25%, 38.50%, and 27.50% higher than the benchmark approaches HSC[9], MDM-CD-RM [16], MDM-ID-RA [16], IDSC [6], SPTC [6], BCF+SVM [50], BCF [50]+1NN, and SIT [48] respectively. As explained in the design of the MSCM method, the proposed sliding chord measures characterize the leaf image not only from its shape, but also from its appearance which results in a richer overall description of the leaf. The existing leaf recognition methods including the state-of-the-art benchmarks only harness the shape features which are effective for species level recognition, but are not suitable for cultivar level classification.

It is interesting to note the effectiveness of the proposed idea of using joint leaf matching for classifying soybean cultivars. This idea is inspired by carefully checking the differences between the leaves from different parts of the same soybean cultivar. From Fig. 4, we found that the leaves from the upper, middle and lower parts of different soybean cultivar plants usually differ in shape and/or appearance. This idea is proven to be effective by our experiments. It can be seen from Table 1 that all the nine competing methods consistently achieved significant improvements on their identification accuracies. The identification accuracies of HSC [9], MDM-CD-RM [16], MDM-ID-RA [16], IDSC [6], SPTC [6], BCF+SVM [50], BCF [50]+1NN, and SIT [48] increased from 14.50% to 34.50%, from 12.25% to 32.00%, from 12.25% to 32.25%, from 13.25% to 33.25%, from 14.50% to 40.25%, from 14.25% to 34.25%, from 12.75% to 29.00%, and

from 15.00% to 40.00%, respectively. While for the proposed method, its identification accuracy improved from 30.50% to an encouraging 67.50%.

To further observe the detailed behaviour of the proposed method in comparison with the benchmark methods in classifying soybean cultivars, we present their intermediate results of the dissimilarity measures in a visualization form. Fig. 6 shows the confusion matrices for the proposed method and the benchmark methods, which are derived from the dissimilarity measure for each pair of samples from different soybean cultivars. Since the version of BCF that uses SVM doesn't provide pairwise similarity, we use the $L_1$-norm distance between the feature vectors of BCF for constructing its confusion matrix. The column of each matrix corresponds to the cultivar index of the first sample and the row of each matrix corresponds to the cultivar index of the second sample in the SoyCultivar leaf database. The dissimilarity measures in the matrices are normalized to the range of [0,1] using its maximum and minimum values and are graphically displayed in gray-scale where black indicates 0 and white indicates 1. It is observed that the diagonal elements (which corresponds to the distance measures between the sample pairs from the same cultivars) of the confusion matrix of the proposed method have obviously smaller values than other elements, significantly better than the benchmark methods.

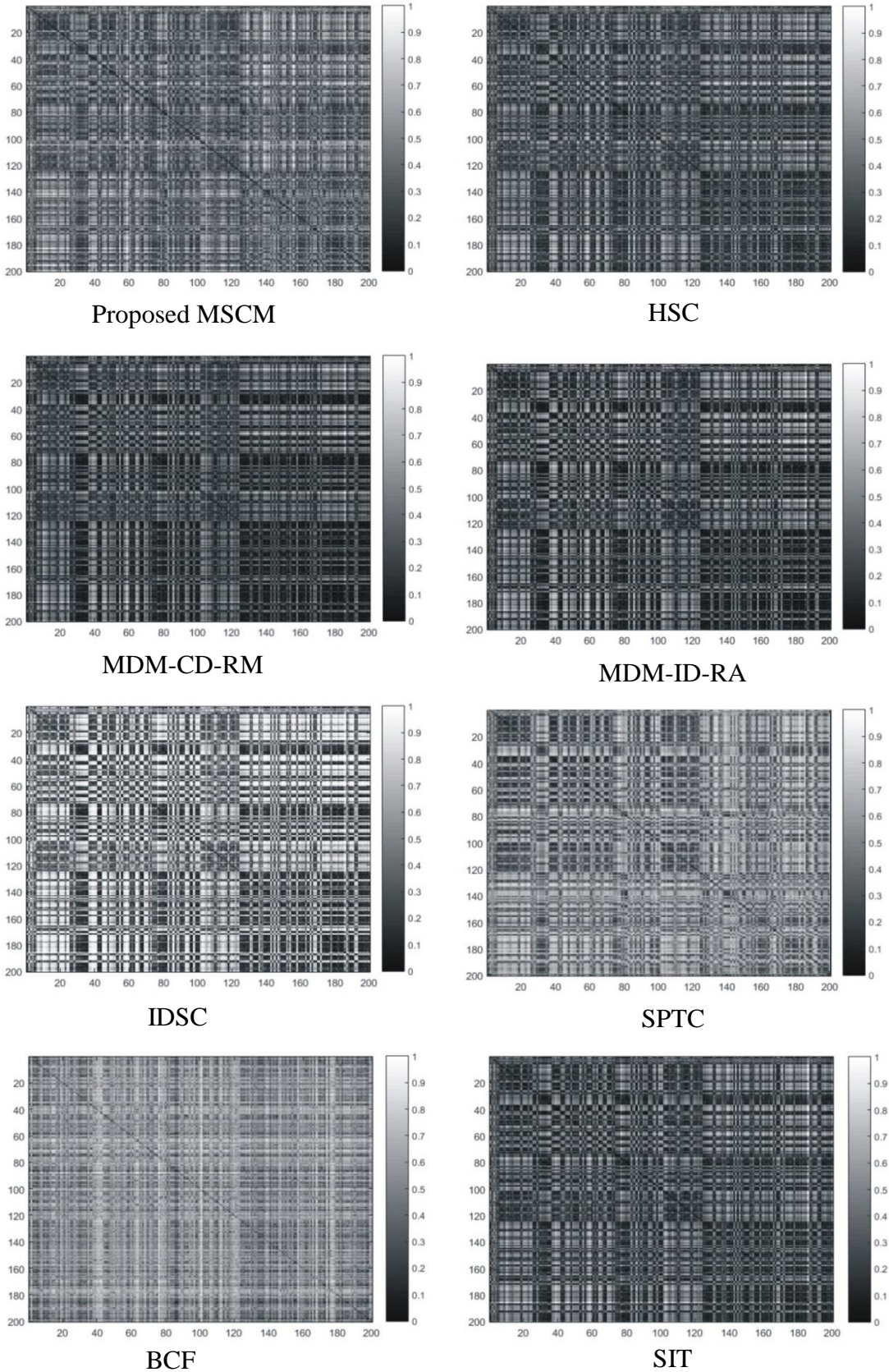

Fig. 6. Confusion matrices obtained by the proposed method and the benchmark approaches that computes the joint leaf dissimilarities. The horizontal axis is the cultivar index of the first sample and the vertical axis is the cultivar index of the second sample in the SoyCultivar database.

To analyse the effect of the weighting factor $W$ in Eq. 16 on the proposed method, a series of experiments are conducted to record the identification accuracy of our algorithm by varying the value of $W$ from 0 to 1 with a step of 0.01, resulting 101 experimental results. These results are plotted as a curve of accuracy versus $W$ (see Fig. 7). It can be seen that when $W = 0$, i.e., only appearance features are used, the proposed method achieves an accuracy of 51%. With the increase of $W$, i.e. the contribution of shape features, the accuracy increases continuously and reaches its peak 67.5% when $W = 0.29$. The accuracy remains stable at its peak value of 67.5% when $W$ ranges from 0.29 to 0.32, then decreases gracefully by further increasing $W$. It is worth noting that only using the shape features (i.e., $W$=1), the proposed method still achieves an accuracy of 40.25%, which is 5.75%, 8.25%, 8.00%, 7.00%, 6.00%, and 11.25% higher than the shape methods of HSC, MDM-CD-RM, MDM-ID-RA, IDSC, BCF+SVM and BCF+1NN, and is slightly better than the SIT method. It is worth stressing that only using the shape features, the proposed method achieved the same accuracy as the SPTC method that uses both shape and texture information. This demonstrates the strong discriminative power of the proposed MSCM in fine-level cultivar characterization and classification.

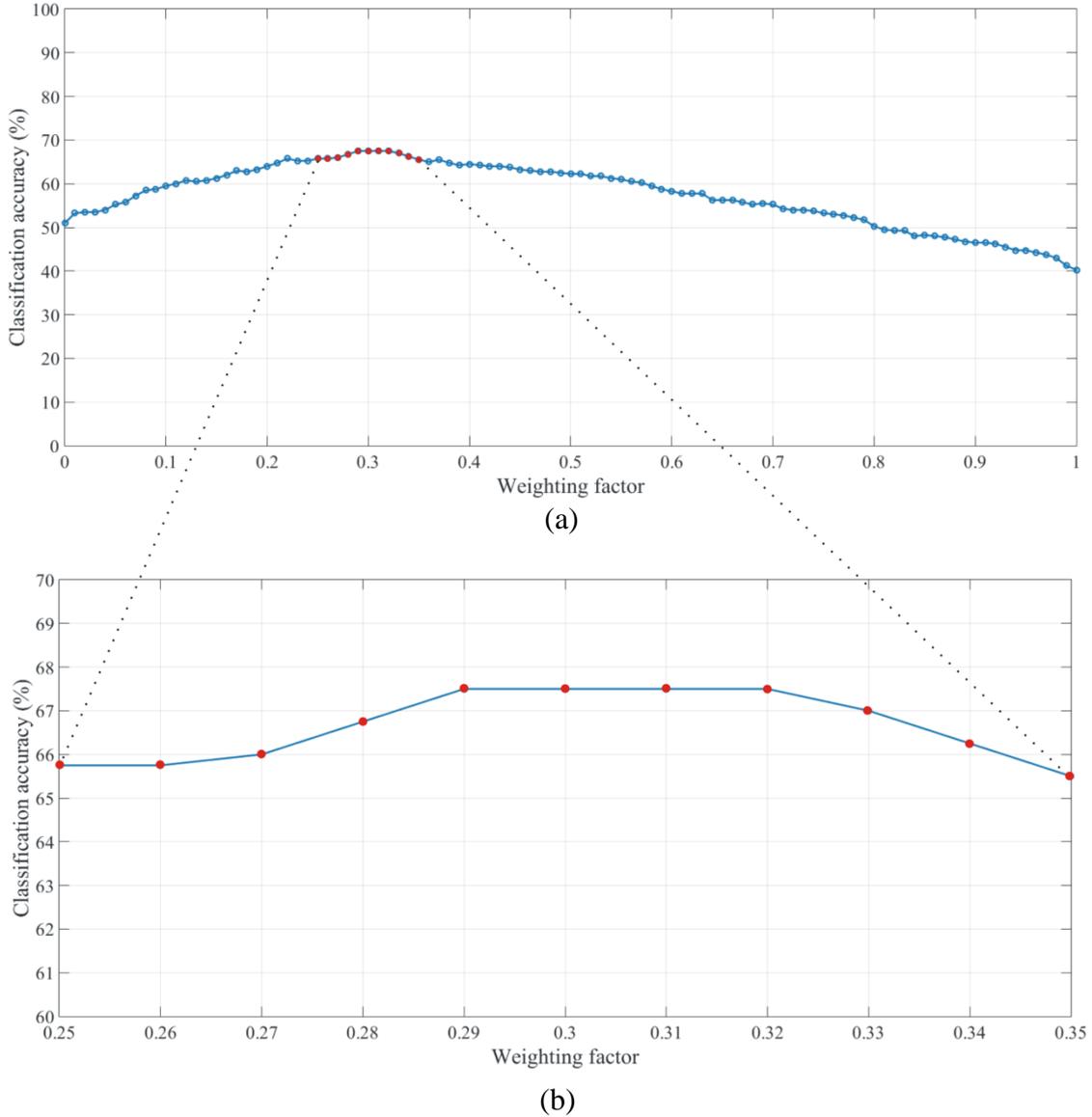

Fig. 7. The identification accuracy of the proposed method versus the weighting factor $W$. (a) The effect of $W$ over its full range. (b) A close look of the plateau area of identification accuracy.

In general, the accuracy of a classification algorithm always degrades when the number of classes in the database increases due to the decrease of average difference between classes. However, a good algorithm is expected to have a relative slower degradation rate. To study the degradation rate of the proposed method with the increase of class number, we conduct another series of experiments by increasing the number of soybean cultivars from 100 to 200 with a step size of 10. The identification accuracies of the proposed method and the eight benchmark methods with an increasing number of cultivars are plotted in Fig. 8. It is encouraging to observe that the proposed method has a slower degradation rate than all the benchmarks methods. When the number of the soybean cultivars increases from 100 to 200, the identification accuracy of the proposed method drops only by 4.50%, lower than the 7.25% drop of the SPTC method that also uses textures, and significantly lower than the over 11% drops of other seven benchmark approaches. This advantage of slow degradation shall be attributed to the fact that the proposed MSCM and SPTC encode both shape and texture information into their descriptors for cultivar classification, which provides them stronger discriminability than the other seven benchmark methods that only use shape information. The above drops of 4.50% (MSCM) vs. 7.25% (SPTC) also demonstrate that the proposed MSCM has stronger discriminative power than the SPTC in capturing the inter-class varieties of cultivar leaf image patterns.

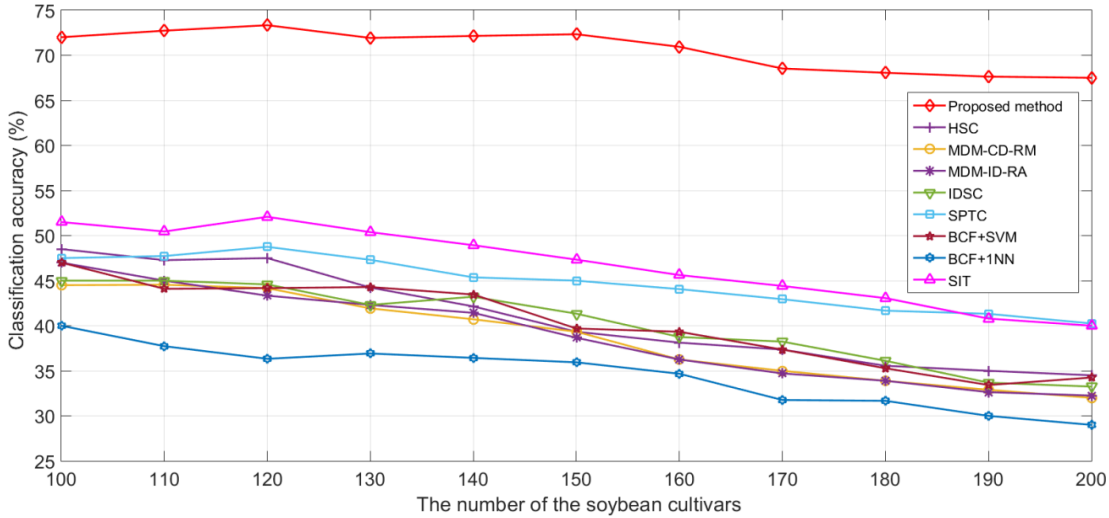

Fig. 8. The curve of the accuracy versus the number of the soybean cultivars.

Recently, Bai et al. [55] proposed that based on any existing shape similarity measure, a new similarity can be learned to improve the accuracy by considering the influences of the neighbours of the given shapes. Their method boosted the discriminability of IDSC [6], one of the benchmarks in this paper, and increased its classification accuracy. Same as in [52], [53], [54], we also applied the learned similarity [55] on the proposed MSCM method and all the benchmark methods for soybean cultivar classification. The identification accuracies boosted by the learned similarity for all the competing methods of using joint matching are summarized in Table 2. It can be seen that the accuracies of all the competing methods increase by a percentage ranging from 1.25% to 11.50%. The proposed method achieves a very encouraging accuracy of 71.75% for the challenging task of cultivar recognition from leaves, which is over 20% higher than the state of the art benchmark methods. The results of this set of experiments indicate the superior performance of the proposed method when further employing a boosting technique of learned similarity.

Table 2. The identification accuracy (%) boosted by the learned similarity [55] for all the competing methods of using joint matching.

| Algorithm | Original | Improved |
|---|---|---|
| HSC [9] | 34.50 | 38.50 |
| MDM-CD-RM [16] | 32.00 | 34.75 |
| MDM-ID-RA [16] | 32.25 | 35.00 |
| IDSC [6] | 33.25 | 39.75 |
| SPTC [6] | 40.25 | 51.75 |
| BCF [50]+1NN | 29.00 | 33.75 |
| SIT [48] | 40.00 | 41.25 |
| **Proposed MSCM** | **67.50** | **71.75** |

# 5. Conclusion

This paper presented a novel multiscale sliding chord matching (MSCM) method for soybean cultivar identification. It can effectively extract the discriminative cultivar features from soybean leaf images in a multiscale sliding chord manner. Chords are geometric elements of the leaf contour and are usually used for capturing properties of an object contour. In this study, we propose to use the chords to regulate integral operations for measuring both the geometrical shape and interior intensity pattern features of soybean leaves in a synchronised way. Multiscale sliding chord measures followed by a joint leaf description of integrating the leaf descriptors from different parts of plants present a powerful characterization for soybean cultivar identification. The effectiveness of the proposed method has been validated on the SoyCultivar leaf database which contains 1200 samples from 200 soybean cultivars. The proposed method achieved exciting identification accuracies that outperformed the state-of-the-art leaf image identification methods.

It is worth noting that two valuable discoveries are made from this research, which may be useful for future soybean cultivar study: (1) The leaves from different parts of soybean plants have different visual cues for soybean cultivar identification and the leaves from the upper part of a plant may carry stronger discriminative power than those from other parts. (2) Joint leaf patterns of integrating the descriptors of leaves from different parts of soybean plants can largely improve the identification accuracy, which may be useful for the study of soybean breeding and cultivation. In future, we will further study the soybean leaf images for more valuable pattern discovery to provide cues for finding the relationship between the genotype and the phenotype of soybean cultivars.

# Appendix A

- $\eta'(t,r) = a \cdot \eta(t,r)$

Proof: For the measure $\eta(t,r)$, its scaled version is

$$\eta'(t,r) = \int_0^{a \cdot l} f'(c'(t,r,\tau'))d\tau' = \int_0^{a \cdot l} f\left(\frac{c'(t,r,\tau')}{a}\right)d\tau' = \int_0^{a \cdot l} f(c(t,r,\tau))d\tau'$$

$$= a \cdot \int_0^l f(c(t,r,\tau))d\tau = a \cdot \eta(t,r). \qquad (17)$$

- $h'(t,r) = a \cdot h(t,r)$

Proof: After the scaling transform, the perpendicular distance $d(t,r,s)$ from the contour point $z(t+s)$ to the chord $L_{t,r}$ (see Eq. (5)) is changed to

$$d'(t,r,s) = \frac{1}{a \cdot l} \left| \det \begin{pmatrix} a \cdot \bar{x}(t) & a \cdot \bar{y}(t) & 1 \\ a \cdot \bar{x}(t+s) & a \cdot \bar{y}(t+s) & 1 \\ a \cdot \bar{x}(t+r) & a \cdot \bar{y}(t+r) & 1 \end{pmatrix} \right|$$

$$= \frac{a^2}{a \cdot l} \left| \det \begin{pmatrix} \bar{x}(t) & \bar{y}(t) & 1 \\ \bar{x}(t+s) & \bar{y}(t+s) & 1 \\ \bar{x}(t+r) & \bar{y}(t+r) & 1 \end{pmatrix} \right| = a \cdot d(t,r,s). \quad (18)$$

Therefore, the measure $h(t,r)$ becomes

$$h'(t,r) = \frac{1}{r} \int_0^r d'(t,r,s) ds = \frac{1}{r} \int_0^r a \cdot d(t,r,s) ds = a \cdot h(t,r). \quad (19)$$

- $\mu'(t,r) = \mu(t,r)$

Proof: For the measure $\mu(t,r)$, its scaled version is

$$\mu'(t,r) = \frac{1}{\eta'(t,r)} \int_0^{a \cdot l} g'(c'(t,r,\tau')) f'(c'(t,r,\tau')) d\tau'$$

$$= \frac{1}{a \cdot \eta(t,r)} \int_0^{a \cdot l} g\left(\frac{c'(t,r,\tau')}{a}\right) f\left(\frac{c'(t,r,\tau')}{a}\right) d\tau'$$

$$= \frac{1}{a \cdot \eta(t,r)} \int_0^{a \cdot l} g(c(t,r,\tau)) f(c(t,r,\tau)) d\tau'$$

$$= \frac{a}{a \cdot \eta(t,r)} \int_0^l g(c(t,r,\tau)) f(c(t,r,\tau)) d\tau = \mu(t,r). \quad (20)$$

- $\sigma'(t,r) = \sigma(t,r)$

Proof: For the measure $\sigma(t,r)$, its scaled version is

$$\sigma'(t,r) = \sqrt{\frac{1}{\eta'(t,r)} \int_0^{a \cdot l} \left(g'(c'(t,r,\tau')) - \mu'(t,r)\right)^2 f'(c'(t,r,\tau')) d\tau'}$$

$$= \sqrt{\frac{1}{a \cdot \eta(t,r)} \int_0^{a \cdot l} \left(g\left(\frac{c'(t,r,\tau')}{a}\right) - \mu(t,r)\right)^2 f\left(\frac{c'(t,r,\tau')}{a}\right) d\tau'}$$

$$= \sqrt{\frac{1}{a \cdot \eta(t,r)} \int_0^{a \cdot l} \left(g(c(t,r,\tau)) - \mu(t,r)\right)^2 f(c(t,r,\tau)) d\tau'}$$

$$= \sqrt{\frac{1}{\eta(t,r)} \int_0^{l} \left(g(c(t,r,\tau)) - \mu(t,r)\right)^2 f(c(t,r,\tau)) d\tau} = \sigma(t,r). \quad (21)$$

# Acknowledgements

This work is supported in part by Australian Research Council (ARC) under Discovery Grants DP140101075.

# References

[1] T.D. Setiyono, A. Weiss, J.E. Specht, K.G. Cassman, A. Dobermann, Leaf area index simulation in soybean grown under near-optimal conditions, Field Crops Research, 108 (2008) 82-92.
[2] J.E. Cavassim, J.C.B Filho, L.F. Alliprandini, R.A.D. Oliveira, E. Daros, E.P. Guerra, AMMI analysis to determine relative maturity groups for the classification of soybean genotypes, Journal of Agronomy, 12 (2013) 168-178.
[3] C.K. Wagner, M.B. McDonald, Rapid laboratory tests useful for differentiation of soybean (Glycine max) cultivars. Seed Sci. Technol. 10 (1982) 431-449.
[4] M.S. Yoon, Q.J. Song, I.Y. Choi, J.E. Specht, D.L. Hyten, P.B. Cregan, BARCSoySNP23: A PANEL OF 23 selected SNPs for soybean cultivar identification, Theor. Appl. Genet., 114 (2007) 885-899.
[5] S. Belongie, J. Malik, J. Puzicha, Shape matching and object recognition using shape contexts, IEEE Trans. Pattern Analysis and Machine Intelligence, 24 (2002) 509-522.
[6] H. Ling, D.W. Jacobs, Shape classification using the inner-distance, IEEE Trans. Pattern Analysis and Machine Intelligence, 29 (2007) 286-299.
[7] P.N. Belhumeur, D. Chen, S. Feiner, D.W. Jacobs, W.J. Kress, H. Ling, I. Lopez, R. Ramamoorthi, S. Sheorey, S. White, L. Zhang, Searching the world's herbaria: A system for visual identication of plant species, ECCV 2008, Part IV, LNCS 5305, 2008, pp. 116-129.
[8] C. Zhao, S.S.F. Chan, W.-K. Cham, L.M. Chu, Plant identification using leaf shapes-A pattern counting approach, Pattern Recognition, 48 (2015) 3203-3215.
[9] B. Wang, Y. Gao, Hierarchical string cuts: a translation, rotation, scale and mirror invariant descriptor for fast shape retrieval, IEEE Trans. Image Processing, 23 (2014) 4101-4111.
[10] A.R. Backes, D. Casanova, O.M. Bruno, A complex network-based approach for boundary shape analysis, Pattern Recognition, 42 (2009) 54-67.
[11] S. Manay, D. Cremers, B.-W. Hong, A.J. Yezzi, S. Soatto, Integral invariants for shape matching, IEEE Trans. Pattern Analysis and Machine Intelligence, 28 (2006) 1602-1618.
[12] H.J. Wolfson, On curve matching, IEEE Trans. Pattern Analysis and Machine Intelligence, 12 (1990) 483-489.

[13] F. Mokhtarian, A.K. Mackworth, A theory of multiscale, curvature-based shape representation for planar curves, IEEE Trans. Pattern Analysis and Machine Intelligence, 14 (1992) 789-805.
[14] T.B. Sebastian, P.N. Klein, B.B. Kimia, On aligning curves, IEEE Trans. Pattern Analysis and Machine Intelligence, 25 (2003) 116-124.
[15] N. Kumar, P.N. Belhumeur, A. Biswas, D.W. Jacobs, W.J. Kress, I. Lopez, J.V.B. Soares, Leafnap: a computer vision system for automatic plant species identification, European Conf. Comput. Vis. (2012) 502-516.
[16] R. Hu, W. Jia, H. Ling, D. Huang, Multiscale distance matrix for fast plant leaf recognition, IEEE Trans. Image Processing, 21 (2012) 4667-4672.
[17] B. Wang, D. Brown, Y. Gao, J. La Salle, MARCH: Multiscale-arch-height description for mobile retrieval of leaf images, Information Sciences, 302 (2015) 132-148.
[18] N. Alajlan, I. EI Rube, M.S. Kamel, G. Freeman, Shape retrieval using triangle-area representation and dynamic space warping, Pattern Recognition, 40 (2007) 1911-1920.
[19] S. Mouine, I. Yahiaoui, A. Verroust-Blondet, A shape-based approach for leaf classification using multiscale triangular representation, in Proceedings of the 3rd ACM International Conference on Multimedia Retrieval, 2013, pp. 127-134.
[20] K. Horaisová, J. Kukal, Leaf classification from binary image via artificial intelligence, Biosystems Engineering, 142 (2016) 83-100.
[21] X.-F. Wang, D.-S. Huang, J.-X. Du, H. Xu, L. Heutte, Classification of plant leaf images with complicated background, Applied Mathematics and Computation, 205 (2008) 916-926.
[22] C.-L. Lee, S.-Y. Chen, Classification of leaf images, International Journal of Imaging Systems and Technology, 16 (2006) 15-23.
[23] M.G. Larese, R. Namías, R.M. Craviotto, M.R. Arango, Automatic classification of legumes using leaf vein image features, Pattern Recognition, 47 (2014) 158-168.
[24] Y. Nam, E. Hwang, D. Kim, A similarity-based leaf image retrieval scheme: Joining shape and venation features, Computer Vision and Image Understanding, 110 (2008) 245-259.
[25] J. Chaki, R. Parekh, S. Bhattacharya, Plant leaf recognition using texture and shape features with neural classifiers, Pattern Recognition Letters, 58 (2015) 61-68.
[26] J.S. Cope, D. Corney, J.Y. Clark, P. Remagnino, P. Wilkin, Plant species identification using digital morphometrics: A review, Expert Systems with Applications, 39 (2012) 7562-7573.
[27] C. Xu, J. Liu, X. Tang, 2D shape matching by contour flexibility, IEEE Trans. Pattern Analysis and Machine Intelligence, 31 (2009) 180-186.
[28] B.W. Hong, S. Soatto, Shape matching using multiscale integral invariants, IEEE Trans. Pattern Analysis and Machine Intelligence, 37 (2015) 151-160.
[29] O. Söderkvist, Computer vision classification of leaves from Swedish trees, Master's thesis, Linköping University, 2001.
[30] P. Novotný, T. Suk, Leaf recognition of woody species in central Europe, Biosystems Engineering, 115 (2013) 444-452.
[31] D. Zhang, G. Lu, Review of shape representation and description techniques, Pattern Recognition, 37 (2004) 1-19.
[32] F. Mokhtarian, A.K. Mackworth, A theory of multi-scale, curvature-based shape representation for planar curves, IEEE Trans. Pattern Analysis and Machine Intelligence, 14 (1992) 789-805.
[33] F. Mokhtarian, S. Abbasi, Matching shapes with self-intersections: application to leaf classification, IEEE Transactions on Image Processing, 13 (2004) 653-661.
[34] Y. Chen, P. Lin, Y. He, Velocity representation method for description of contour shape and the classfication of weed leaf images, Biosystems Engineering, 109 (2011) 186-195.
[35] S. Manay, D. Cremers, B.-W. Hong, A. Yezzi, S. Soatto, Integral invariants for shape matching, IEEE Trans. Pattern Analysis and Machine Intelligence, 28 (2006) 1602-1618.
[36] D. Casanova, J.J. de Mesquita Sá Junior, O.M. Bruno, Plant leaf identification using Gabor wavelets, International Journal of Imaging Systems and Technology, 19 (2009) 236-243.
[37] J.S. Cope, P. Remagnino, S. Barman, P. Wilkin, Plant texture classification using Gabor co-occurrences, In: International Symposium on Visual Computing, Las Vegas, Nevada, USA, Springer-Verlag, 2010, pp. 669-677.
[38] A.R. Backes, D. Casanova, O.M. Bruno, Plant leaf identifcation based on volumetric fractal dimension, Int J. Pattern Recognit. Artif. Intell. 23 (2009) 1145-1160.
[39] M.A.J. Ghasab, S. Khamis, F. Mohammad, H.J. Fariman, Feature decision-making ant colony optimization system for an automated recognition of plant species, Expert Systems with Applications, 42 (2015) 2361-2370.

[40] S. Venkatesh, R. Raghavendra, Local gabor phase quantization scheme for robust leaf classification, In: 2011 Third National Conference on Computer Vsion, Pattern Recognton, Image Processing and Graphics (NCVPRIPG), pp. 211-214.
[41] J.K. Park, E.J. Hwang, Y. Nam, Utilizing venation features for efficient leaf image retrieval, J. Syst. Softw. 81 (2008) 71-82.
[42] O.M. Bruno, R. de Oliveira Plotze, M. Falvo, M. de Castro, Fractal dimension applied to plant identification, Information Sciences, 178 (2008) 2722-2733.
[43] D.J. Hearn, Shape analysis for the automated identifcation of plants from images of leaves, Taxon, 58 (2009): 934-954.
[44] C. Meade, J. Parnell, Multivariate analysis of leaf shape patterns in Asian species of the Uvaria group (Annonaceae), Botanical Journal of The Linnean Society, 143 (2003): 231-242.
[45] R. de Oliveira Plotze, M. Falvo, J.G. Pádua, L.C. Bernacci, M.L.C. Vieira, G.C.X. Oliveira, O.M. Bruno, Leaf shape analysis using the multiscale Minkowski fractal dimension, a new morphometric method: a study with Passiflora (Passifloraceae), Candian Journal of Botany, 83 (2005) 287-301.
[46] M. Hasegawa, S. Tabbone, Amplitude log Radon transform for geometric invariant shape descriptor, Pattern Recognition, 47 (2014) 643-658.
[47] N. Nacereddine, S. Tabbone, D. Ziou, Similarity transformation parameters recovery based Radon transform. Application in image registration and object recognition, Pattern Recognition, 48 (2015) 2227-2240.
[48] B. Wang, Y. Gao, Structure integral transform versus Radon transform: A 2D mathematical tool for invariant shape recognition, IEEE Transcations on Image Processing, 25 (2016) 5635-5648.
[49] B.-W. Hong, S. Soatto, Shape mathching using multiscale integral invariants, IEEE Trans. Pattern Analysis and Machine Intelligence, 37 (2015) 151-160.
[50] X. Wang, B. Feng, X. Bai, W. Liu, L.J. Latecki, Bag of contour fragments for robust shape classification, Pattern Recognition, 47 (2014) 2116-2125.
[51] S.H. Lee, C.S. Chan, S.J. Mayo, P. Remagnino, How deep learning extracts and learns leaf features for plant classification, Pattern Recognition, 71 (2017) 1-13.
[52] R. Hu, W. Jia, Y. Zhao, J. Gui, Perceptually motivated morphological strategies for shape retrieval, Pattern Recognition, 45 (2012) 3222-3230.
[53] V. Premachandran, R. Kakarala, Perceptually motivated shape context which uses shape interiors, Pattern Recognition, 46 (2013) 2092-2102.
[54] R.A. Guler, S. Tari, G. Unal, Landmarks inside the shape: Shape matching using image descriptors, Pattern Recognition, 49 (2016) 79-88.
[55] X. Bai, X. Yang, L.J. Latecki, W. Liu, Z. Tu, Learning context-sensitive shape similarity by graph transduction, IEEE Trans. Pattern Analysis and Machine Intelligence, 32 (2010) 861-874.